\documentclass[review]{elsarticle}

\usepackage{lineno,hyperref}
\usepackage{amsmath}
\usepackage{amssymb}
\usepackage{mathtools}
\usepackage{cancel}
\usepackage{float}
\usepackage{multirow}
\usepackage{xcolor}
\usepackage{enumitem}
\usepackage{algorithm}
\usepackage{algpseudocode}
\usepackage{tabularx}
\usepackage{caption}

\journal{Arxiv}

\begin{document}

\begin{frontmatter}

\title{Multi-fidelity reinforcement learning framework for shape optimization}

\author{Sahil Bhola\corref{mycorrespondingauthor}}
\address{Department of Aerospace Engineering, University of Michigan, Ann Arbor, MI 48109, USA}
\cortext[mycorrespondingauthor]{Corresponding author}
\ead{sbhola@umich.edu}

\author{Suraj Pawar}
\address{School of Mechanical and Aerospace Engineering, Oklahoma State University, Stillwater, OK 74078, USA}

\author{Prasanna Balaprakash}
\address{Mathematics and Computer Science Division, Argonne National Laboratory, Lemont, IL 60439, USA}

\author{Romit Maulik}
\address{Mathematics and Computer Science Division, Argonne National Laboratory, Lemont, IL 60439, USA}

\begin{abstract}
Deep reinforcement learning (DRL) is a promising outer-loop intelligence paradigm which can deploy problem solving strategies for complex tasks. Consequently, DRL has been utilized for several scientific applications, specifically in cases where classical optimization or control methods are limited. One key limitation of conventional DRL methods is their episode-hungry nature which proves to be a bottleneck for tasks which involve costly evaluations of a numerical forward model. In this article, we address this limitation of DRL by introducing a controlled transfer learning framework that leverages a multi-fidelity simulation setting. Our strategy is deployed for an airfoil shape optimization problem at high Reynolds numbers, where our framework can learn an optimal policy for generating efficient airfoil shapes by gathering knowledge from multi-fidelity environments and reduces computational costs by over 30\%. Furthermore, our formulation promotes policy exploration and generalization to new environments, thereby preventing over-fitting to data from solely one fidelity. Our results demonstrate this framework's applicability to other scientific DRL scenarios where multi-fidelity environments can be used for policy learning.
\end{abstract}


\end{frontmatter}


\section{Introduction}
Shape optimization techniques have been of critical interest to the scientific
community, with applications in structural design~\cite{Ding1986, Haftka1986, Hsu1994}
, electromagnetics~\cite{Kim2009, Lalau-Keraly2013, Mori2015, Akcelik2005}, medicine~\cite{Cassioli2013}, and energy sciences~\cite{Zhao2013}. For e.g.,
within the aerospace industry such optimization techniques led to design explorations at unforeseen configurations~\cite{Mohammadi2004, nemec2004multipoint, samareh2001survey, kenway2016multipoint}. Gradient-based optimization methods such as adjoint-based scale well with problem complexity, however, often result in sub-optimal design as they get trapped in local minima~\cite{Skinner2018}. Gradient-free methods (e.g. particle swarm, genetic algorithm) are easy to implement but require large number of samples and have weaker optimality criterion~\cite{Skinner2018}. For shape optimization problems in complex fluid flows which are inherently high-dimensional and non-linear such algorithms
often under-perform due to multiple local minima and high computational cost~\cite{Skinner2018}. Over recent years, DRL has gained traction in the fluid dynamics community for learning non-linear decision-making policies for complex tasks in an unsupervised manner~\cite{Garnier2021, Rabault2020}.
Applications include flow sculpting~\cite{Lee2018}, active flow control at low Reynolds number~\cite{Ma2018, Tang2020} and
under turbulent conditions~\cite{Ren2020}, controlling chaos~\cite{Bucci2019, Vashishtha2020}, accelerating model
convergence~\cite{Pawar2021}, turbulence modeling~\cite{Novati2021}, reduced order
modeling~\cite{Benosman2020, Bassenne2019}, controlling instability~\cite{Belus2019}, and controlling
Rayleigh-Benard convection~\cite{Beintema2020}.
Reinforcement learning offers a framework for learning decision-making policies based on successive interactions with the computational
model (\textit{environment}). In DRL an \textit{agent} (neural network)
observes some \textit{state} (partial or noisy) of the system and predicts an \textit{action} to be executed on the environment in
order to receive some heuristic driven \textit{reward}~\cite{Sutton2018}. The goal of the learning algorithm is to learn
a policy, \(\pi\)($action\mid state$) that maximizes the cumulative reward. Since design exploration is based on trials (\textit{episodes}), these
methods perform well with problems having multiple local minima or environmental
uncertainty. However, for shape optimization problem in fluid mechanics learning design policies directly from high-fidelity environment through such trial evaluations can get prohibitively expensive. This is primarily due to the prohibitive cost of executing each action coupled with the episode-hungry nature of DRL. For e.g. Viquerat \textit{et al.}~\cite{Viquerat2021} reported using approx. 4000 such episodes for learning the airfoil design policy. In the present work we introduce controlled transfer learning (CTL) framework which can exploit multi-fidelity environments for policy learning and mitigates negative transfer. We demonstrate the performance of our framework by considering an airfoil shape optimization problem at high Reynolds number.

To the authors' knowledge there have been limited investigations of shape optimization in fluid
mechanics using DRL. Q-learning based methods (see~\S\ref{sec:Methodology}) could predict the airfoil
shape for a range of flight conditions~\cite{Lampton2008a, Lampton2009}. However, these methods did not scale with increasing
action (shape parameterization) and state space, requiring coarse grid approximations.
Novati \textit{et al.}~\cite{Novati2017} examined two self-propelled swimmers to learn the optimal motion
pattern to reduce energy expenditure using deep Q-learning with experience replay~\cite{Mnih2015}.
However, to make computations tractable these simulations used Reynolds average Navier Stokes (RANS)
to model the environment. A similar study was performed by Verma \textit{et al.}~\cite{Verma2018} using
high-fidelity direct numerical simulations (DNS) along with LSTMs to model the
environment and parameterize the policy, respectively. Using such high-fidelity models
reduced the model-form uncertainty but increased the computational cost of learning
the policy. More recently, Viquerat \textit{et al.}~\cite{Viquerat2021} used a degenerate
multi-environment reinforcement learning~\cite{Rabault2019, Ghraieb2021} framework to directly predict the airfoil shape using
the Proximal Policy Optimization algorithm (see~\S\ref{sec:Methodology}). Despite improved scalability due to parallel learning the cost of computing the model
(RANS solver) resulted in sub-optimal learning (4000 episodes were required). It was reported that reward shaping may accelerate learning and thereby reduce some computational cost. Despite such
improvements the large computational overhead of evaluating the high-fidelity environment
inhibits the use of DRL for fluid mechanics problems. Thus, to learn policies derived from high-fidelity environments it is crucial to develop frameworks that scale well with increased model complexity.

Transfer Learning (TL) offers a framework to incorporate the learning experience from one task (source task) and use it to improve the learning for a similar task (target task)~\cite{Taylor2009}. Yan \textit{et. al.}~\cite{Yan2019} used transfer learning to utilize the learning experience from semi-empirical model to guide missile shape design policy learned using RANS models. Improvement in design policy and a drastic reduction in computational time was reported in their study. Li \textit{et. al.}~\cite{Li2021a} carried a similar study for the design of super-critical airfoils using DRL. A low-fidelity surrogate model was constructed using sparse high-fidelity evaluations, and was used to learn the initial design policy, i.e. the source task. Similarly, Li et.al.~\cite{Li2021} used Gaussian processes to create a low-fidelity environment for source task training. These studies show that by leveraging multi-fidelity environments we can accelerate policy learning on the target task, improve asymptotic performance of the agent, and significantly reduce the high-fidelity evaluations required for DRL. This requires a careful selection of the source task which is similar to the target task and is also tractable to learn. Despite of considering source task selection these studies did not address \textit{optimal source task learning} required for optimal knowledge transfer to the target task~\cite{Taylor2009}. Excessive learning on the source task could inhibit generalizability and policy exploration on the target task and thereby result in a sub-optimal target task policy. In this article the introduced framework CTL addresses optimal source task learning thereby improving target task policy learning. To summarize, this article comprises of the following key contributions,
\begin{enumerate}
    \item A novel CTL framework is introduced that (a) can exploit multi-fidelity environments for scalable learning in DRL, and (b) addresses optimal source task learning to prevent negative transfer. 
    \item We demonstrate the performance of our framework for an airfoil shape optimization problem at high Reynolds number (similar to Viquerat \textit{et al.}~\cite{Viquerat2021}). 
\end{enumerate}

The manuscript is organized as follows. In~\S\ref{sec:Methodology} we introduce the reinforcement learning framework and the multi-fidelity formulation used for controlled transfer learning. Subsequently, in~\S\ref{sec:Results_and_Discussion} we present the results obtained for learning airfoil shape optimization policy using our framework. We provide our conclusions in~\S\ref{sec:Conclusion}.


\section{Methodology}
\label{sec:Methodology}

\subsection{Reinforcement learning}
\label{subsec:Reinforcemet_learning}

In reinforcement learning the \textit{agent} ingests information about the \textit{environment} in the form of a \textit{state} ($s_{t}$) and proposes an \textit{action} ($a_{t}$). This action is executed on the environment to advance it to the next state 
$s_{t+1}$. The agent receives a \textit{reward} ($r_{t}$) from the environment which is typically a heuristic 
measure of the quality of the action taken. Finally, the objective of the learning algorithm is to learn a policy, $\pi(a \vert s) \in \Pi:  s\in S \rightarrow a\in A$. The policy is a function that receives the state of the system as an input and returns an action, i.e., $\pi(a \vert s)=\text{p}(a_t=a \vert s_t=s)$. The goal of an RL agent is to maximize the future rewards defined as follows
\begin{equation}
    G_{t} \triangleq \sum_{k=t} ^{T}\gamma^{k}r_{k},
    \label{eq:reward}
\end{equation}
where $\gamma \in [0, 1]$ is the discount factor and $T$ is the episode length. The discount factor weighs the importance between the short-term and long-term benefits of taking an action. The formulation can be represented as a Markov Decision Process (MDP) \cite{Sutton2018} 
resulting in a state-action-reward trajectory \( \tau = \{s_{t}, a_{t}, r_{t}, s_{t+1}, a_{t+1}, r_{t+1}, \dots s_{T}, a_{T}, r_{T}\}\).  The RL agent is trained to find a policy that optimizes the expected return when starting in the state $s_t$ at time step $t$ and is called as V-value function and it can be written as follows
\begin{equation}
    V^{\pi}(s) \triangleq \mathbb{E}_\pi \left[ G_t | s_t, \pi \right]
\end{equation}
where $G_t$ is the reward function defined in equation~(\ref{eq:reward}).

Similarly, the expected return starting in a state $s_t$, taking an action $a_t$, and thereafter following a policy $\pi$ is called as the Q-value function and can be written as
\begin{equation}
    Q^\pi(s_t, a_t) \triangleq \mathbb{E}_\pi\left[G_t | s_t, a_t, \pi\right]
    \label{eqn:Q_function}
\end{equation}
Optimal decision-making policy can be learned via value-based methods \cite{szepesvari1999unified} or policy-based methods \cite{arulkumaran2017deep}. 
Value-based methods train agents that maximize the $Q$-value function.
However, for high-dimensional action spaces, evaluating the $Q$-value function becomes intractable. Policy-based methods learn the parameterized policy distribution $\pi_{\theta}(a\mid s)$, enabling us to sample actions from a continuous distribution, $p(a\mid s)$. In the following, we briefly describe our environment and reward metrics, and then explain the reinforcement learning formulation.

\subsection{Computational environment}
\label{subsec:CFD_environments}

In this work, our motivation is to perform goal-oriented aerodynamic shape optimization with an aim to minimize drag. Specifically, we consider an airfoil shape parameterized using B\'ezier curves. The B\'ezier curve is defined using six control points and the leading-edge radius, i.e., it has thirteen design variables. The flow around the airfoil can be modeled with hierarchies of models ranging from Panel methods to direct numerical simulation. Typically, the flow dynamics is governed by a non-dimensional parameter Reynolds number based on chord length and the inflow speed.

An optimal design policy is learned by shaping the reward function such that the desired objective of minimizing the drag is achieved. Thus, we consider the (state, action, reward) triplet: Reynolds number based on chord length ($Re_{c}$), airfoil shape, and the negative drag coefficient ($-C_{d}$). We use the degenerate Deep RL framework~\cite{Viquerat2021} with $T = 1$ which means that the complete simulation of the flow solver corresponds to one episode.
For each episode, the environment (i.e., two-dimensional flow solver) observes a fixed state sample according to $Re_{c}\sim\mathcal{N}(\mu, \sigma)$.
The action of the agent is to perturb the shape of the airfoil by changing the airfoil design variables, i.e., $A\in\mathbb{R}^{13}$. To illustrate the performance of the CTL framework we have considered two environments/computational models: potential flow solver and RANS solver.
Here, potential flow solver is the low-fidelity environment and RANS is the high-fidelity environment.

Here, we briefly describe the RANS solver which is utilized as the high-fidelity environment
for interacting with the RL agent. The RANS solver is a steady-state solver,
and it numerically solves the time-averaged Navier-Stokes equations. The RANS equations
are derived using Reynolds decomposition, where an instantaneous quantity is decomposed
into time-averaged and fluctuating quantities. The RANS equations for incompressible
flow can be expressed as follows

\begin{align}
    \frac{\partial \overline{u}_i}{\partial x_i} &=0 \\
    \rho \overline{u}_j\frac{\partial \overline{u}_i}{\partial x_j} &= \rho \overline{f}_i + \frac{\partial}{\partial x_j}\bigg[-\overline{p}\delta_{ij} + 2\mu \overline{S}_{ij} - \rho \overline{u_i^\prime u_j^\prime} \bigg]
\end{align}
where $(\overline{\cdot})$ denoted time-averaged quantities, $u$ is velocity component, $u^\prime$ is the fluctuating component, $p$ is the pressure, $\rho$ is the density of fluid, $f$ represents
external forces, $\mu$ is the dynamic viscosity, and $\overline{S}_{ij}$ is the mean strain-rate tensor. The non-linearity of the Navier-Stokes equations introduces the term $(\rho \overline{u_i^\prime u_j^\prime})$ and is referred to as the Reynolds stress. The computation of Reynolds stress requires some modeling and is usually done through an additional algebraic or differential equation. In this study, we adopt the one-equation Spalart-Allmaras model \cite{spalart1992one} as the
turbulence closure model for RANS equations.

\begin{figure}
    \centering
    \includegraphics[width = 0.5\textwidth]{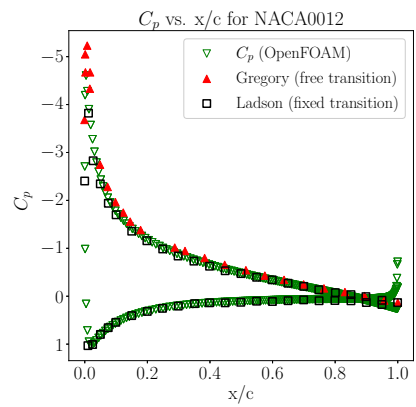}
    \caption{Distribution of pressure coefficient along the chord length of the NACA0012
    airfoil for Reynolds number $Re=3e+6$ and angle of attack $\alpha=10^{\circ}$.
    The experimental data for validation are obtained from Gregory \textit{et. al.}~\cite{Gregory1970} and Ladson \textit{et. al.}~\cite{Ladson1988}.}
    \label{fig:cfd_validation}
\end{figure}

All computational simulations are performed with the SimpleFOAM solver in OpenFOAM. SimpleFOAM is based on the finite-volume discretization for RANS equations, and it follows a SIMPLE algorithm to conjugate pressure field and velocity field. The C-grid domain is utilized such that the outlet boundary is located at $20C$ ($C = 1$ is the chord length of the airfoil) from the trailing edge of an airfoil and the total height of the computational domain is $20C$. The height of
the first cell near the airfoil surface is set at $0.001C$ and the boundary layer growth rate is fixed at 1.05. This mesh resolution within the boundary layer is found to be adequate to resolve the turbulent boundary layer region with sufficient accuracy. Fig.~\ref{fig:cfd_validation} shows the pressure coefficient along the chord length of NACA0012 airfoil compared with the experimental results from the literature~\cite{Gregory1970, Ladson1988}. We can see that there is a good match between the results of our CFD simulations and the experimental data.

\subsection{Learning framework}
\label{subsec:Learning_framework}

In DRL, a neural network is utilized as an RL agent, and therefore, the weights and biases of the neural network are the parameters of the policy~\cite{Sutton2000}. The policy parameterized by $\theta \in \mathbb{R}^d$ is denoted as $\pi_\theta(\cdot)$ and the agent is trained with an objective function $J(\theta)$ defined as follows
\begin{equation}
    J(\theta) \triangleq V^{\pi_\theta}(s_0);\quad \label{eq:obj_function}
\end{equation}
where an episode starts in some initial state $s_0$, and $V^{\pi_\theta}$ is the value function for the policy $\pi_\theta$. The parameters of the neural network $\theta$ are learned by estimating the gradient of an objective function with respect to trainable parameters and updating them using a gradient ascent algorithm as follows
\begin{equation}
    \theta \leftarrow \theta + \alpha \nabla_\theta J(\theta),    \label{eq:gradient_ascent}
\end{equation}
where $\alpha$ is the learning rate of the optimization algorithm. The gradient of an objective function can be computed using the policy gradient theorem \cite{Sutton2000} as follows
\begin{eqnarray}
    \nabla_\theta V^{\pi_\theta}(s_0) = \mathbb{E}_{\pi_\theta}\big[ \nabla_\theta \big(\log ~\pi_\theta(s,a) \big) Q^{\pi_\theta}(s,a)].
\end{eqnarray}

There are two main challenges in using the above empirical expectation. The first one is the substantial number of samples required and the second is the difficulty of obtaining stable and steady improvement. There are various categories of policy-gradient algorithms that are proposed to address these issues such as actor-critic algorithm \cite{Konda2000} and trust region policy optimization (TRPO) \cite{Schulman2015a}. In the present work, we use the Deep Proximal Policy Optimization (PPO) \cite{Schulman2017} algorithm which belongs to the class of policy-based methods. The PPO is motivated by the same question as TRPO, i.e., how to achieve the biggest possible improvement step on a policy without taking the large step that can cause performance collapse. The performance of policy gradient methods is extremely sensitive to the learning rate $\alpha$ in equation~(\ref{eq:gradient_ascent}). If the learning rate is too large it can cause the training to be unstable. While TRPO tries to solve this problem with a second-order method, the PPO algorithm uses a clipped surrogate objective function to avoid excessive updates in policy parameters in a simplified way. The clipped objective function of the PPO algorithm is  
\begin{multline}
    J^{{clip}}(\theta) = \mathbb{E}\big[ {min}(r_t(\theta) {A}^{\pi_\theta}(s,a), {clip} (r_t(\theta),1-\epsilon,1+\epsilon) 
    {A}^{{\pi_\theta}}(s,a) ) \big], \label{eq:ppo_clip}
\end{multline}
where $r_t(\theta)$ denotes the probability ratio between new and old policies as follows
\begin{equation}
    r_t(\theta) \triangleq \frac{\pi_{\theta+\Delta \theta}(s,a)}{\pi_{\theta}(s,a)}.
\end{equation}
 The advantage function ${A}^{\pi_\theta}$ in equation~(\ref{eq:ppo_clip}) is defined as follows 
\begin{equation}
    A^{\pi}(s,a) \triangleq Q^\pi(s,a) - V^\pi(s).
\end{equation}
The $\epsilon$ in equation~(\ref{eq:ppo_clip}) is a hyperparameter that controls how much a new policy is allowed to be deviated from the old. This is done using the function ${clip}(r_t(\theta),1-\epsilon,1+\epsilon)$ that enforces the ratio between new and old policy ($r_t(\theta)$) to stay between the limit $[1-\epsilon,1+\epsilon]$. 

\begin{figure}[t!]
    \centering
    \includegraphics[width = 1\textwidth]{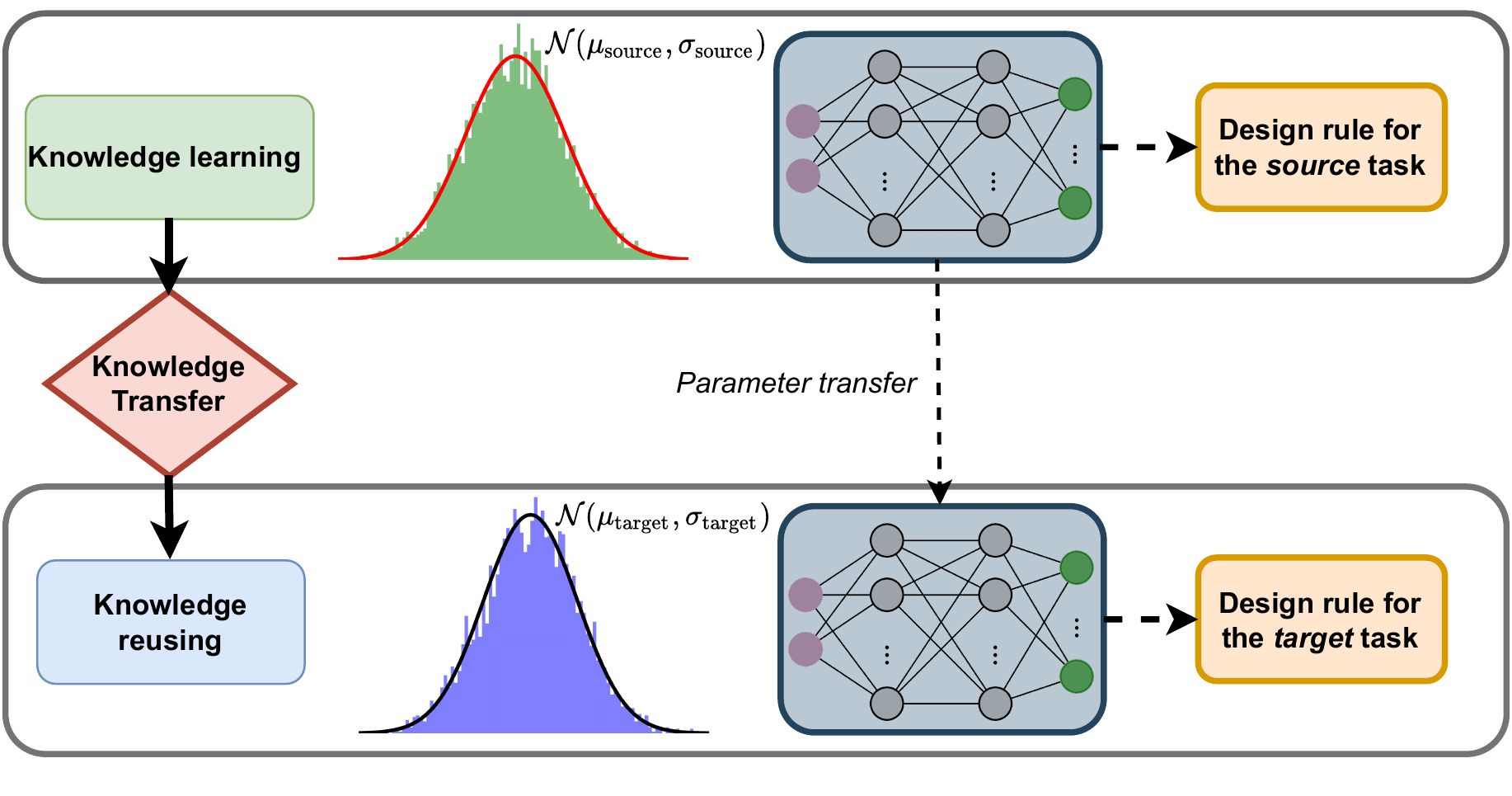}
    \caption{Multi-fidelity reinforcement learning framework.}
    \label{fig:learning_framework}
\end{figure}

Figure~\ref{fig:learning_framework} illustrates the CTL
framework used in this study. In the \textit{knowledge learning} phase of the algorithm the agent observes states according to a source distribution, $(Re_{c})_{source}\sim\mathcal{N}(\mu_{source}, \sigma_{source})$ and interacts with an inexpensively evaluated low-fidelity environment. This enables the agent to learn the policy for the source task in a cost-efficient manner. We introduce a variance ratio ($\beta_{e}$) for transfer control that prevents over-fitting on the source task and promotes generalization. This variance ratio is defined as

\begin{equation}
    \beta_{e} \triangleq \frac{{\xi_{e}}}{max({\xi_{e}},{\xi_{e-1}},\dots
    {\xi_{e}}
    )},
    \label{eq:variance_red_ratio}
\end{equation}
where $\xi_{\Tilde{e}} = \{ r_{\Tilde{e}}, r_{\Tilde{e}-1}, \dots r_{\Tilde{e}-k}\}$ with a look-back window of size $k$, and $e\in\{1, \cdots, episode_{max}\}$. By evaluating the rewards over a look-back window we exploit the locality of reward convergence. This avoids delay in knowledge transfer due to slow convergence of the reward function on the source task. 
Our choice of the structure of the variance ratio ensures that as the agent is learning the ratio will decrease, as the variance in the rewards will decrease. We initialize by considering $\beta_{1} = 1$ and for $e<k$ we evaluate the denominator in equation~(\ref{eq:variance_red_ratio}) over the available reward values, i.e., consider $\Tilde{k}<k$. Introducing such a transfer control criterion in the learning framework ensures that agent extracts \textit{sufficient information} from the source task. We define a cut-off percentage $\Gamma = 0.3$ such that when $\beta_{e}<\Gamma$ we label the knowledge learning task as complete. Algorithm 1 summarizes the CTL algorithm for task learning. Once the agent has gathered sufficient knowledge about the environment on the source task and developed a policy for it, we transfer the agent to the target task with $(Re_{c})_{target}\sim\mathcal{N}(\mu_{target}, \sigma_{target})$ in the \textit{knowledge reusing} phase. In the present work we transfer knowledge by sharing the parameters of the deep neural network which parameterizes the policy~\cite{Taylor2009}.

\begin{algorithm}
\textbf{Controlled transfer learning}
\begin{algorithmic}
\Require $\quad k\neq 0$, $\Gamma$, $episode_{max}$
\Ensure $\quad \beta_{e} = 1$
\State e=2
\While{e $<episode_{max}$}
\If{e$<$k}
\State Compute $\beta_{e}$ via equation~(\ref{eq:variance_red_ratio}) with k $\gets$ e
\Else
\State Compute $\beta_{e}$ via equation~(\ref{eq:variance_red_ratio})
\EndIf
\If{$\beta_{e}\leq \Gamma$}
\State Learning policy for the task complete
\EndIf
\EndWhile
\end{algorithmic}
\end{algorithm}

The target environment can be a low (or high) fidelity environment with a different state distribution. We consider \textit{single-fidelity} CTL and \textit{multi-fidelity} CTL which have a low-fidelity and a high-fidelity target environment, respectively. These studies encapsulate that the policies learning on the target task can be improved by (a) learning on a source task with a similar state distribution at the same fidelity, or (b) utilizing prior knowledge from a lower fidelity (in-expensive) source task. Specifically, for the latter, policy learned from a low-fidelity environment is often sub-optimal and can be improved by transferring to a higher fidelity environment. Thus, by examining single-fidelity and multi-fidelity CTL, we decouple the effects of experience gained from a similar state distribution and by higher fidelity evaluations. Besides using such multi-fidelity environments, we use a distributed learning~\cite{Pawar2021} framework which parallelizes the RL framework itself. By distributing the learning over processors and pooling the experience we can further accelerate learning.


\begin{figure}[h!]
    \centering
    \includegraphics[width=1\textwidth]{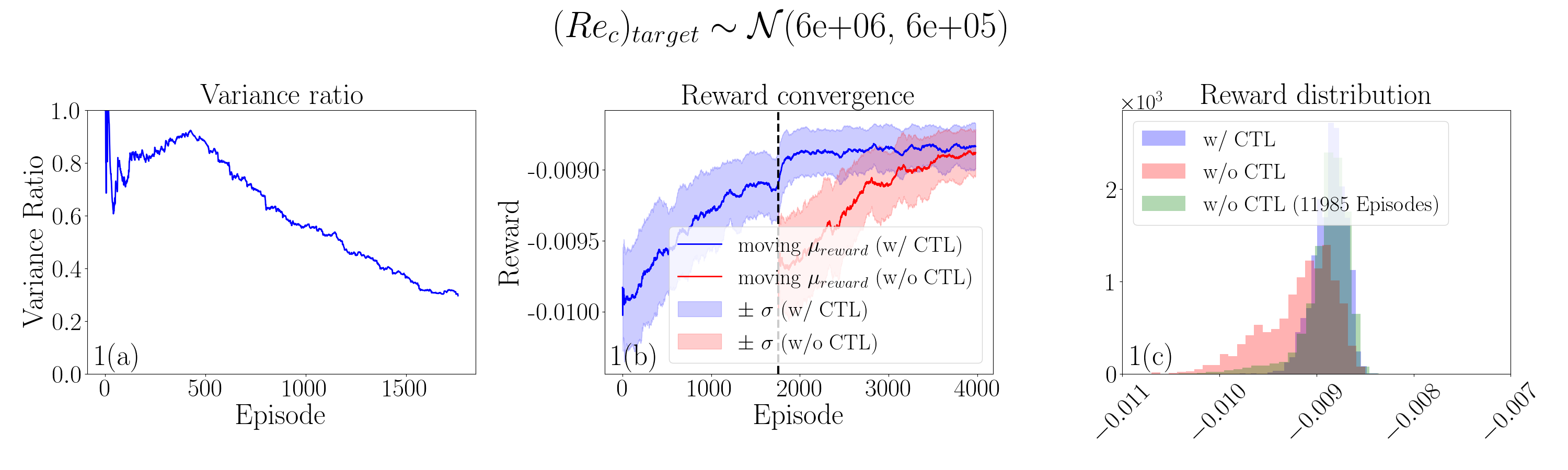}
    \includegraphics[width=1\textwidth]{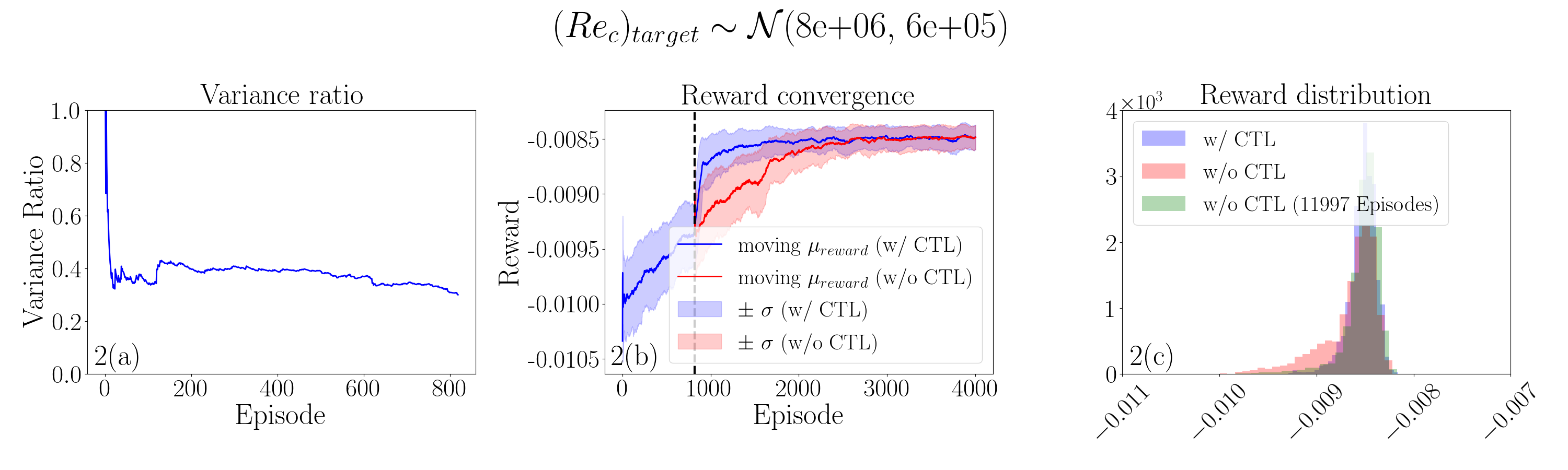}    \includegraphics[width=1\textwidth]{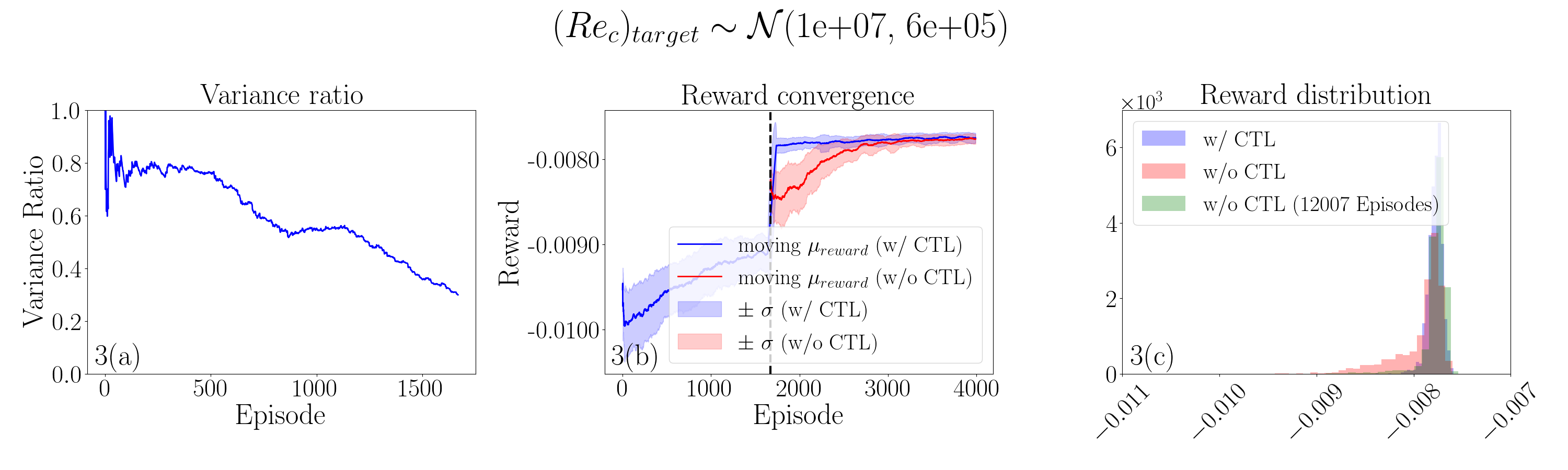}
    \caption{Variance ratio convergence (left column), reward convergence (center column), and reward distribution using last 500 episodes (right column) for single-fidelity CTL with $(\mu_{source} = 5.5e+6, \sigma_{source}=5e+5)$ and
    $\mu_{target}$ = $6e+6$  (top row), $8e+6$ (middle row) and $1e+7$ (bottom row). $\sigma_{target}=5e+5$ for all cases.}
    \label{fig:Campaign1/Intra_model_TL}
\end{figure}

\section{Results and Discussion}
\label{sec:Results_and_Discussion}

In this section we examine the single-fidelity and multi-fidelity CTL framework by considering $(Re_{c})_{source}\sim\mathcal{N}(5.5e+6, 5e+5)$ for all the cases.  
We consider three different target state boundary conditions: $\mu_{target}\in$ \{6e+6, 8e+6, 1e+7\} with $\sigma_{target}=5e+5$ for all our cases. For similar learning tasks where mutual information between the actions from the source task and target task, i.e., $I(a_{source}\sim p_{source}(a\mid s);a_{target}\sim p_{target}(a\mid s)) \neq 0$, the agent will exploit the policy learned from the source task to guide policy exploration on the target task. On the other end of the spectrum, when $I(a_{source}\sim p_{source}(a\mid s);a_{target}\sim p_{target}(a\mid s)) = 0$, policy learning without any knowledge transfer will be recovered for the target task. Although, we have not considered such information-theoretic measures to show task similarity, it is intuitive to consider two airfoil shape optimization tasks as similar~\cite{Taylor2009}.

For the distributed learning framework, we have used $T_{L}=20$ as often used in several studies\cite{Rabault2019, Rabault2019a} with 4 cores. This means that actions will be individually sampled according to the learned policy and executed on different processors. The experience will be accumulated after $T_{L}$ episodes and will be used to update the policy.

\subsection{Single-fidelity CTL}
\label{subsec:Intra_model_TL}
We begin by training the agent on a low-fidelity source environment with $(Re_{c})_{source}\sim\mathcal{N}(5.5e+6, 5e+5)$ followed by knowledge transfer to the target task (also a low fidelity computational environment) with state distribution $(Re_{c})_{target}\sim\mathcal{N}(\mu_{target}, \sigma_{target})$. Figure~\ref{fig:Campaign1/Intra_model_TL} illustrates the evolution of variance ratio, reward and the obtained reward
distribution for $\mu_{target}$ = $6e+6$, $8e+6$, and $1e+7$ with and without CTL. As discussed in~\S\ref{subsec:Learning_framework} the variance ratio will decrease as the policy is being learned for the source task (see frames 1(a), 2(a) and 3(a)). However, as expected this convergence is not monotonic because the policy is being learned in a stochastic manner. By using variance ratio for transfer control, i.e., a source task convergence criterion we identify local reward convergence.

Despite considering target state distributions that were around $8\sigma_{source}$ away from $\mu_{source}$ the transferred agent quickly adapts to the unfamiliar environment (see frames 1(b),
2(b) and 3(b)). As observed and expected, the agent adapts much faster for target state distributions that are closer to the source state distributions (frame 1(b)). Nevertheless, we were able to obtain a significant acceleration in learning and a
computational saving of 1000-2000 episodes. Through CTL we were able to obtain less variance in our reward prediction for the same number of training episodes (see frames 1(c), 2(c) and 3(c)). Further, we compare reward distributions between CTL and the converged distributions obtained on the target task. Here, to obtain the converged distribution we train our model for the target task for sufficiently substantial number of episodes. We notice that in addition to outperforming the case without CTL (for same training episodes) we are much closer to the converged reward distribution.

\begin{figure}[t]
    \centering
    \includegraphics[width=1\textwidth]{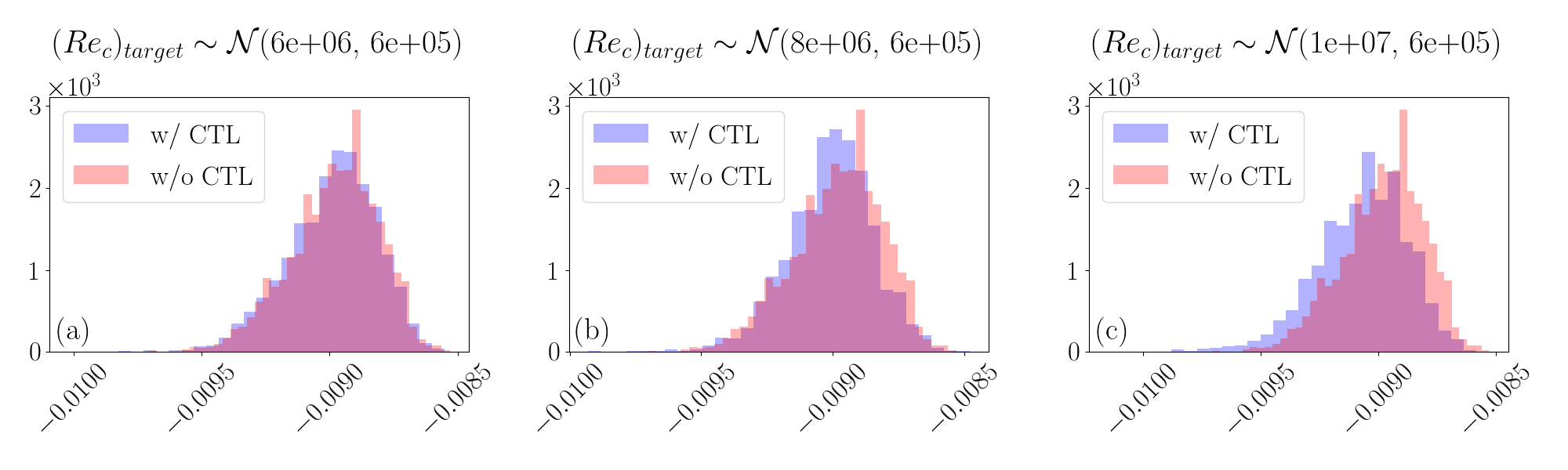}
    \caption{Reward distribution obtained by an agent trained on the target distribution (with CTL) tested on the source distribution $(U_{source}\sim\mathcal{N}(5.5e+6, 5e+5)$) for 5000 episodes.} 
    \label{fig:source_dist_testing}
\end{figure}

Figure~\ref{fig:source_dist_testing} shows the reward distribution obtained by an agent trained on the target distribution (with CTL) \textit{tested} on the source distribution. We compare our results with the reward distribution obtained
by testing an agent trained (for 10000 episodes) on the source distribution. For all the cases the reward distribution agrees well, however, we notice that for $\mu_{target}=1e+7$ the performance of the agent deteriorates (see frame (c)). This is expected since the agent observes very few states that belong to the high
probability region of the source state distribution, and as the training on the target distribution progresses the policy `forgets' such states.

\subsection{Multi-fidelity CTL}
\label{subsec:Inter_model_TL}

\begin{figure}
    \centering
    \includegraphics[width=1\textwidth]{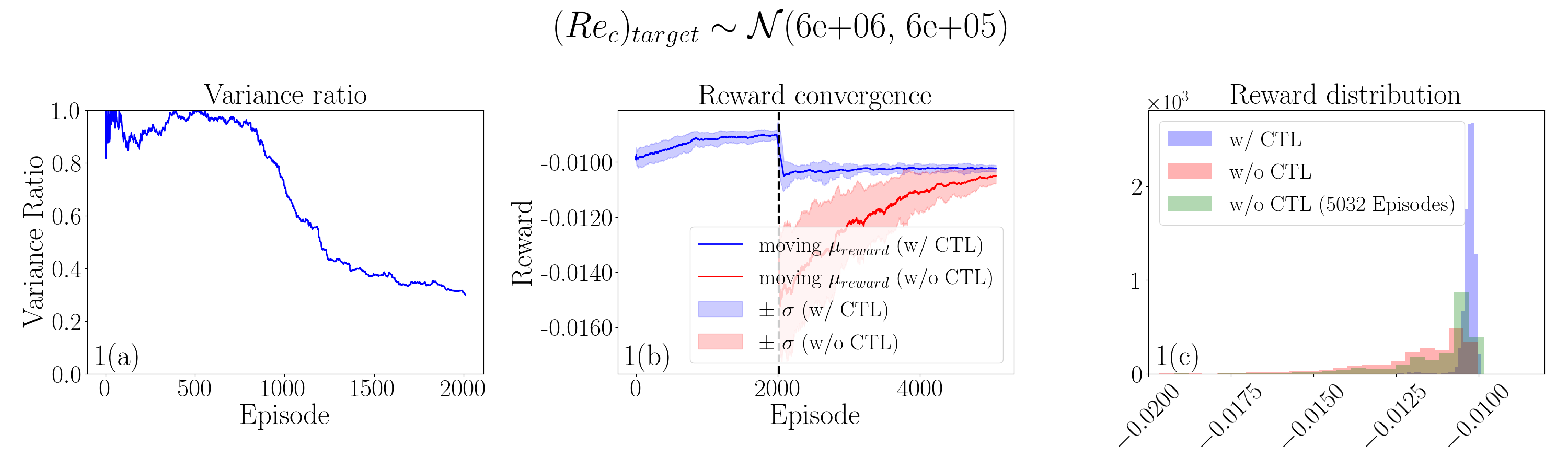}
    \includegraphics[width=1\textwidth]{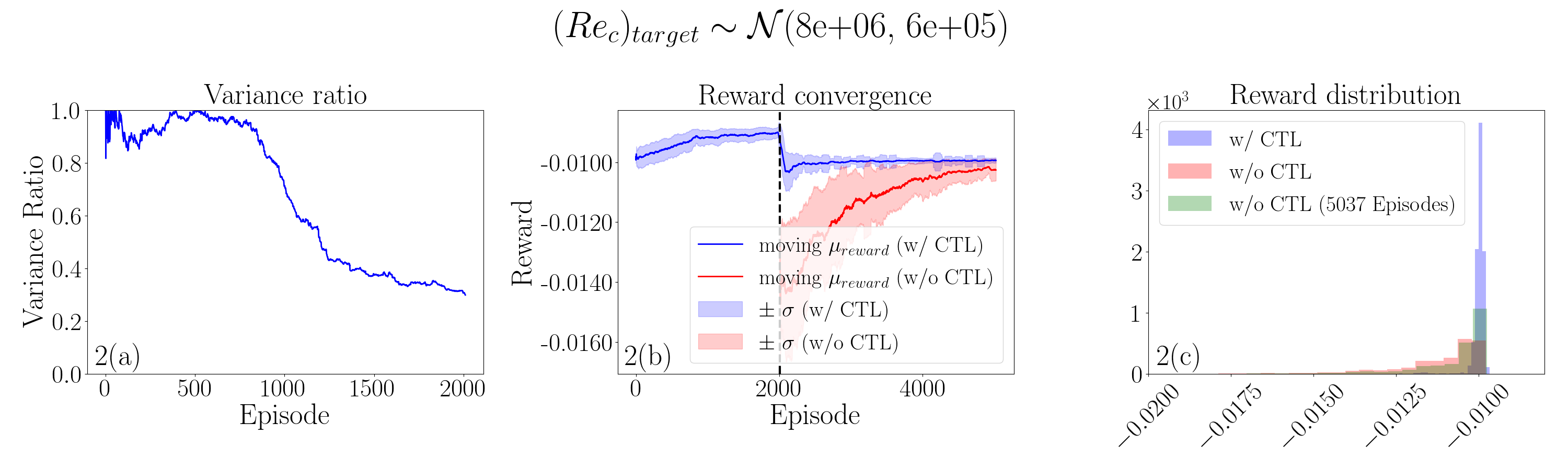}    \includegraphics[width=1\textwidth]{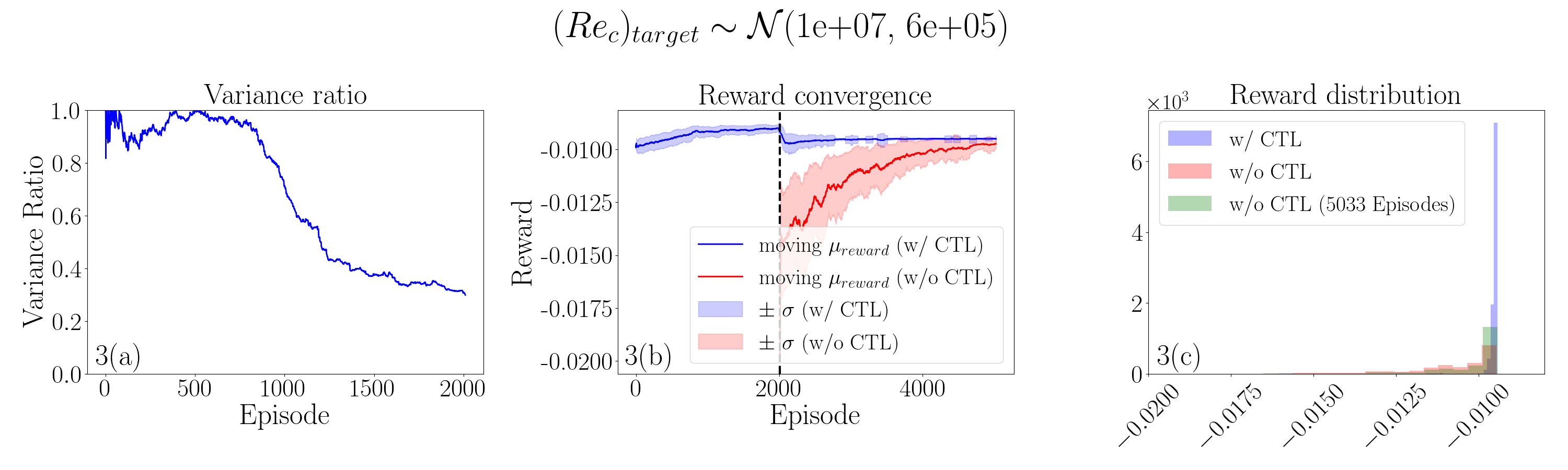}
    \caption{Variance ratio convergence (left column), reward convergence (center column), and reward distribution using last 500 episodes (right column) for multi-fidelity CTL with $(\mu_{source} = 5.5e+6, \sigma_{source}=5e+5)$ and
    $\mu_{target}$ = $6e+6$ (top row), $8e+6$ (middle row) and $1e+7$ (bottom row). $\sigma_{target}=5e+5$ for all cases.}
    \label{fig:Campaign2/Inter_model_TL}
\end{figure}

We now examine the model performance by training the agent on a high-fidelity target
environment with prior knowledge of the policy from a similar low fidelity environment.
The idea is to exploit the low computational cost of the low fidelity environment
to learn a crude policy, then refine the policy for the target task using a
high fidelity environment. Figure~\ref{fig:Campaign2/Inter_model_TL} illustrates the
variance ratio convergence, reward convergence, and the obtained reward
distribution for the multi-fidelity CTL. Like the single-fidelity CTL we
notice that the agent adapts to the target task quickly, reducing the episode evaluation
on the high-fidelity model by over 30\%. This mitigates the enormous computational
cost of finding high fidelity optimal design using RL. Moreover, as observed in the
single-fidelity CTL the model produces less variation in the obtained rewards than without
CTL.

\begin{figure}[ht!]
    \centering
    \includegraphics[width= 1\textwidth]{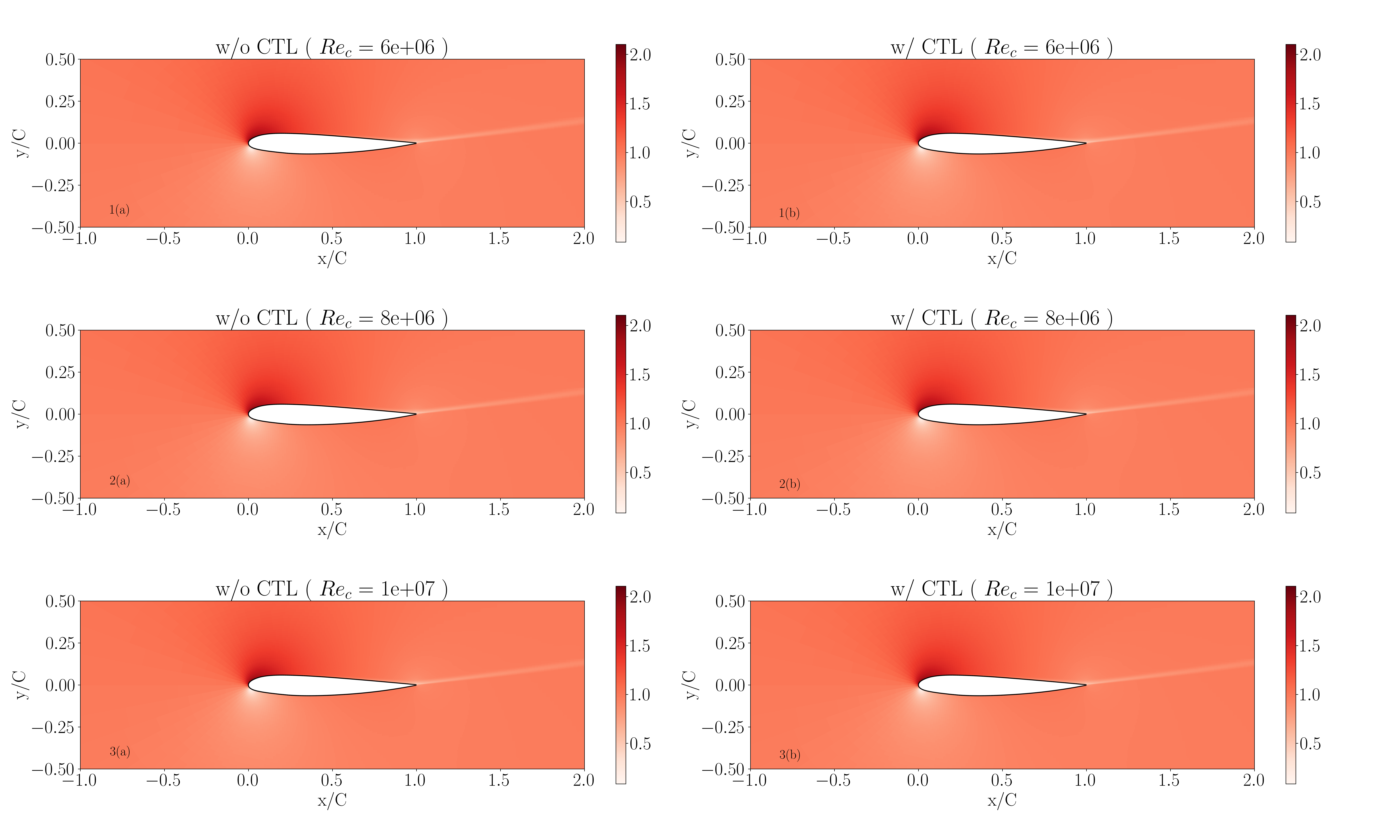}
    \caption{Non dimensional velocity field obtained for the mean predictive airfoils without CTL (Left column) and with CTL (Right column). Velocity is non-dimensionalized with mean free-stream velocity according to $Re_{c}$}
    \label{fig:velocity_field}
\end{figure}

\begin{table}[ht!]
    \centering
    \begin{tabular}{|c|c|c|}
    \hline
        $Re_{c}$ & Case & $C_{d}$ \\
    \hline
    \multirow{2}{2.5em}{6e+6}     & w/o CTL & 0.0102\\
                            & w/ CTL & 0.0102\\
  \hline
    \multirow{2}{2.5em}{8e+6}    & w/o CTL & 0.0099\\
                            & w/ CTL & 0.0099\\
    \hline
    \multirow{2}{2.5em}{1e+7}    & w/o CTL & 0.0094\\
                            & w/ CTL & 0.0094\\
    \hline
    \end{tabular}
    \caption{$C_{d}$ values obtained using mean predicted shape at $Re_{c}$ = $\{6e+6, 8e+6, 1e+7\}$.}
    \label{tab:drag}
\end{table}

\begin{figure}
    \centering
    \includegraphics[width = 1\textwidth]{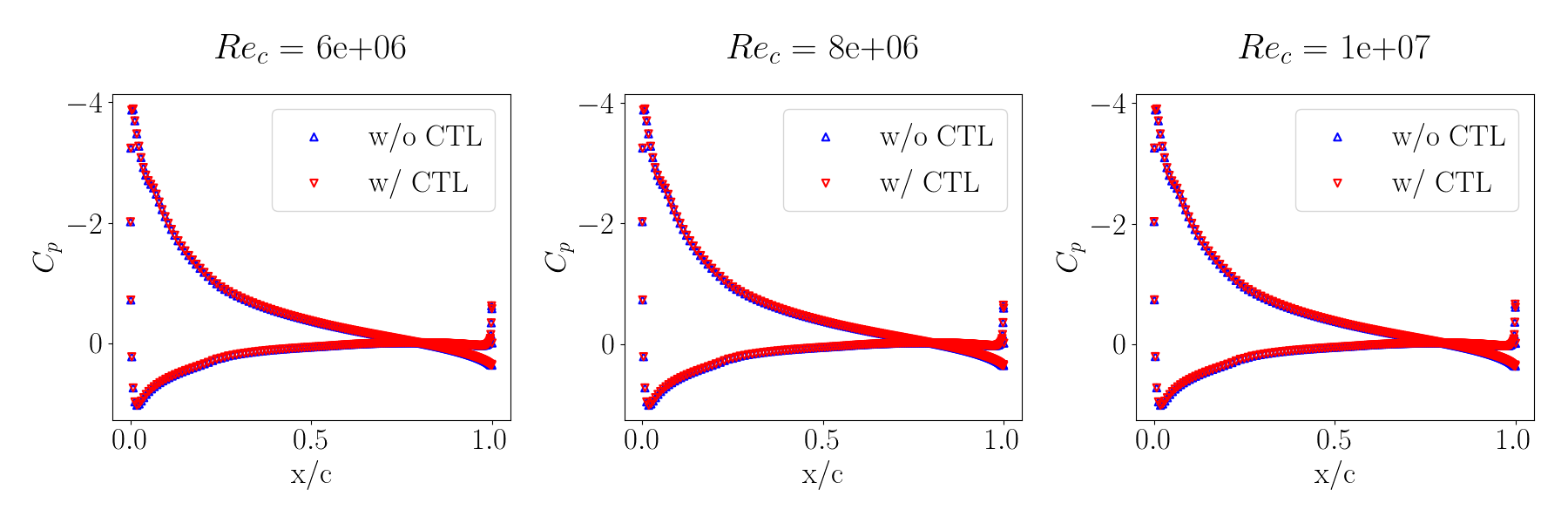}
    \caption{$C_{p}$ vs. $x/c$ at Reynolds number, $Re_{c}$ = $6e+6$ (Left), $8e+6$ (Center), and $1e+7$ (Right) using mean predictive airfoil with and without CTL.}
    \label{fig:pressure_dist}
\end{figure}
    
Using the mean predictive shape obtained from the target task we examine the velocity
field for a fixed $Re_{c} \in \{6e+6, 8e+6, 1e+7\}$ as illustrated in figure~\ref{fig:velocity_field}.
Qualitatively, obtained velocity field with multi-fidelity CTL agrees with the field
obtained without CTL. Moreover, as tabulated in table~\ref{tab:drag} the obtained
drag coefficient ($C_{d}$) is the same with and without CTL. Despite such good agreement, it is to be noted that the parameterization of the action space affects the learned airfoil shape. Increasing the dimensionality of the action space could result in noticeable
differences between the predicted airfoil shape and thereby the obtained drag. Moreover, as shown in figure~\ref{fig:pressure_dist} the predicted pressure distribution using multi-fidelity CTL agrees well with the pressure distribution over the
mean predictive airfoils obtained using just high-fidelity model for DRL training.

\section{Conclusion}
\label{sec:Conclusion}
In this article, we have introduced a novel controlled transfer learning (CTL) framework for multi-fidelity reinforcement learning to address the episode-hungry nature of policy-based reinforcement learning methods. We demonstrate this framework for aerodynamic optimization, which requires several expensive computational evaluations of a forward model and is thus intractable with conventional variants of policy-based methods. Our framework relies on transfer learning between policies derived from computational environments of varying fidelities. In addition, we introduce control transfer learning to dynamically determine the appropriate number of episodes required for policy learning at a particular fidelity before transfer is initiated. We assess this framework for two types of scenarios. The first scenario corresponds to one where the transfer learning is performed between models that are trained for the same computational fidelity but for different boundary conditions (single-fidelity CTL). This experiment demonstrates how our framework can adapt a previously obtained policy for a novel environment. The second scenario is one where the environment is generated from a computational environment which is significantly more expensive (multi-fidelity CTL). Here, we demonstrate that our framework provides significant computational savings by leveraging the learning from prior low-fidelity experiences. Specifically, we are able to reduce the computational cost (measured by the number of high-fidelity evaluations) by over 30\%. Therefore, our framework allows for the generalization of a learned policy to new computational environments without requiring retraining from scratch. Extensions to this work will revolve around information-theoretic techniques for assessing policy transfers between concurrent reinforcement learning tasks. This is particularly interesting for scalable asynchronous reinforcement learning where models at various fidelities provide computational environments of varying complexities for heterogeneous architectures.

\section*{Acknowledgements}

This material is based upon work supported by the U.S. Department of Energy (DOE), Office of Science, Office of Advanced Scientific Computing Research, under Contract DE-AC02-06CH11357. This research was funded in part and used resources of the Argonne Leadership Computing Facility, which is a DOE Office of Science User Facility supported under Contract DE-AC02-06CH11357.

\section*{Data availability}

The data and code related to the experiments performed in this article are available at \url{https://github.com/Romit-Maulik/PAR-RL}.

\bibliography{references}

\end{document}